\documentclass[sigconf]{acmart}
\renewcommand\footnotetextcopyrightpermission[1]{} % removes footnote with conference information in first column
\usepackage{tabulary}
\usepackage{booktabs}
\usepackage{fancyhdr}
\AtBeginDocument{%
  \providecommand\BibTeX{{%
    \normalfont B\kern-0.5em{\scshape i\kern-0.25em b}\kern-0.8em\TeX}}}

\acmPrice{15.00}
\acmISBN{978-1-4503-XXXX-X/18/06}

\setcopyright{none}
\begin{document}

\title{Application of BERT in Wind Power Forecasting-Teletraan's Solution in Baidu KDD Cup 2022}


\author{Longxing Tan}
\affiliation{%
  \institution{Teletraan Inc.}
  \city{Hangzhou}
  \country{China}}
\email{lx.tan@teletraan.io}

\author{Hongying Yue}
\affiliation{%
  \institution{Renmin University of China}
  \city{Beijing}
  \country{China}}
\email{yuehongyingyhy@163.com}



\begin{abstract}
  Nowadays, wind energy has drawn increasing attention as its important role in carbon neutrality and sustainable development. When wind power is integrated into the power grid, precise forecasting is necessary for the sustainability and security of the system. However, the unpredictable nature and long sequence prediction make it especially challenging. In this technical report, we introduce the BERT model applied for Baidu KDD Cup 2022, and the daily fluctuation is added by post-processing to make the predicted results in line with daily periodicity. Our solution achieves 3rd place of 2490 teams. The code is released at \href{https://github.com/LongxingTan/KDD2022-Baidu}{github.com/LongxingTan/KDD2022-Baidu}

\end{abstract}






\maketitle

\section{Introduction}
Wind energy plays an important role in carbon neutrality. The precise prediction of wind power can help its integration into the power system with sustainability and security of supply
\cite{hanifi2020critical}. However, the unpredictable nature of wind makes it challenging, especially for long sequence forecasting.

Transformer models are proposed for long sequence forecasting in recent research \cite{wen2022transformers}. To utilize the advantage of its capturing long dependencies, we implement a BERT model for wind power future prediction. Our solution is a single model with a single BERT block, with both accuracy and efficiency.

\subsection{Dataset}
A unique spatial, dynamic wind power forecasting dataset-SDWPF \cite{zhou2022sdwpf} is provided by Baidu and Longyuan for this competition.
The training data consists of 245-day records for 134 turbines. Its interval is 10 minutes. Additionally, each sample has wind speed, wind direction, relative power, the temperature, in total ten parameters. The procedures of data preprocessing are introduced in Section 4.1.

\subsection{Task}
Given at most 14-day historical data, participants need to predict the future 2 days' wind power for every turbine and every 10 minutes. There are 142 samples provided in the last phase to evaluate the model performance, and each sample is randomly sampled over several months. Inference on the test data needs to be finished in 10 hours in the online environment.

\subsection{Evaluation}
The performance is measured by the average of RMSE (root mean square error) and MAE (mean absolute error) for each turbine as metrics score. There are some invalid values in ground truth labels such as zero values, missing values, and unknown values. Based on the task description, the metrics score for those invalid values in ground truth is set to zero to ignore them. 

\section{Related Work}
There is already a long history of wind power forecasting. The mainstream approaches include the physical approach, statistical approach and the ensemble approach \cite{kariniotakis2004state}. The physical approach uses the meteorological service and numerical weather prediction to predict the future in a short. 

The statistical approach builds a model using statistical methods. The commonly used time series prediction methods include ARIMA, statistical models\cite{cao2018multi}, and deep learning methods \cite{swaminathan2021wind}. Besides, now the long sequence time series forecasting also catches lots of attention from the real-world requirements. Recently, the prediction capacity is increased by the latest transformer family, such as the Informer \cite{zhou2021informer} and Autoformer \cite{wu2021autoformer} model. Informer uses the sparse attention and one-time decoder to accelerate the inference. The Autoformer uses a decomposition method and frequency domain attention to improve its capacity in long sequence prediction. The spatial-temporal approach is also a promising way for wind power forecasting \cite{li2022deep}. The spatial information can be used to improve the forecasting accuracy \cite{fang2020constgat}. So batches of new models are proposed and achieve good results with the development of graph neural networks, like GCN-LSTM, Graph Wavenet \cite{yu2017spatio}, Graph Convolutional Network \cite{stanczyk2021deep} and Graph Attention Network \cite{zhou2020distance}. 

\section{Solution Overview}
In this section, we explain the main components of our solution. The overall architecture of our method is shown in Figure 1.

\begin{figure*}[h]
  \centering
  \includegraphics[width=\linewidth]{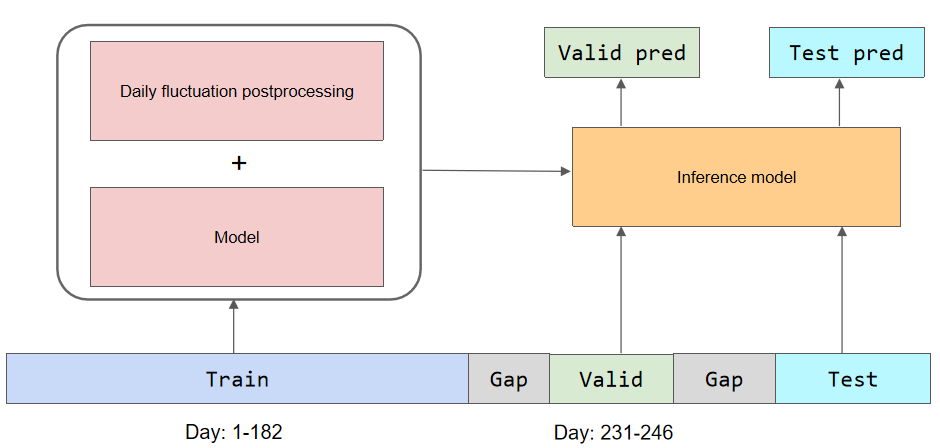}
  \caption{Solution architecture}
  \Description{Solution architecture}
\end{figure*}

We implement two pipelines to treat the problem either as a multi-step time series prediction or a spatial-temporal prediction. What's more, we can choose to predict the generated wind power directly or to predict the wind speed first and then transfer it into wind power\cite{wang2021review}. Besides we tried training different time scale models to handle the long sequence prediction.

When it comes to a time series task, each turbine can generate its samples independently. Because the whole wind farm has a very similar trend, the time series model can capture the long-term seasonality better for long sequence prediction. We tried KNN, LightGBM, RNN, TCN, BERT, Seq2seq, Wavenet, and Transformer models. 

When it comes to a spatial-temporal task, the promising models are like GCN-LSTM, ConvLSTM \cite{shi2015convolutional}, GraphWavenet \cite{wu2019graph}, Graph Transformer \cite{yun2019graph} and spatial CNN models. If spatial information and spatial distribution could be captured, it's assumed to be helpful for future prediction as well \cite{wang2021libcity}.

\section{Detailed method}
\subsection{Data preprocessing}

We create the samples for each turbine every 10 minutes by sliding the window to get more training data. There are NANs and invalid values in the data. We use the corresponding previous value to fill them for training and inference. And we use a min-max scaler to standardize the features into scope between zero and one, but leave the target column alone.

\subsection{Features}
To avoid over-fitting, we try to use as fewer features as possible. The commonly used lag features, rolling features, and time features are not helpful according to our experiments. The temperature is noisy, so the temperature feature is not involved. 

Given the unpredictable nature, it's hard to predict the future trend and seasonality perfectly. So we try to add spatial information and temporal information as features to improve the prediction performance. But they are not helpful in our experiments. So finally we only choose the wind speed and wind direction to train the model.

\subsection{Models}
We choose a single BERT model \cite{devlin2018bert} as our final model to handle the long dependencies for long sequence prediction. The BERT model has been widely used in all tasks in natural language processing since its publication. In computer vision, speech, and time series, it has also been widely investigated and it achieves fruitful results since then. The detailed model structure we use is shown in Figure 2.

\begin{figure*}[h]
  \centering
  \includegraphics[width=\linewidth]{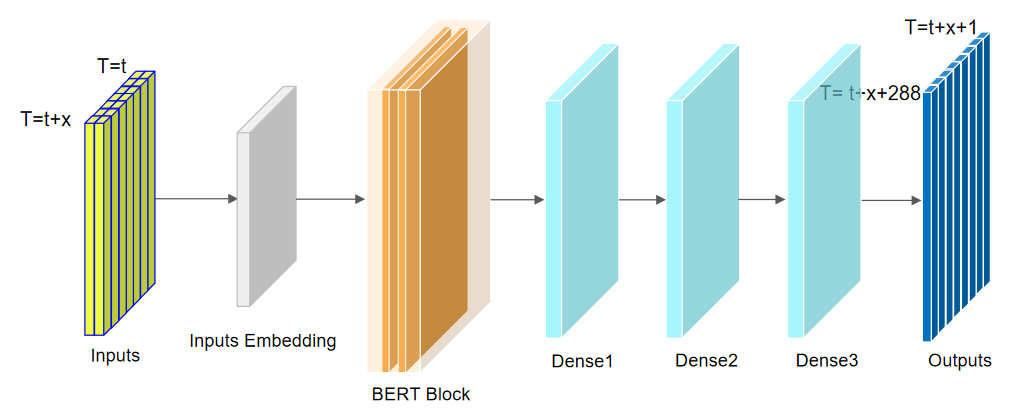}
  \caption{BERT model structure}
  \Description{BERT model structure}
\end{figure*}

A token embedding layer without positional encoding can project the raw feature space into the attention space. The experiments show there is not a significant influence having positional encoding or not. The module following the embedding layer is a standard BERT encoder, with self-attention layer, feed-forward network, layer normalization and residual connection. Then 3 dense layers with dropout can capture the crossed information from time and features. The hidden sizes of the last dense layer are equal to the predicted steps 288. And our experiments show that layer normalization and residual connection are both important that can't be ignored. The detailed model configuration is shown in Table 1.

\begin{table}
  \caption{BERT model configuration}
  \label{tab:freq}
  \begin{tabular}{ccl}
    \toprule
    Hyper-parameters&Value\\
    \midrule    
        Train sequence lengths & 288 \\ 
        Predict sequence lengths & 288 \\ 
        Number of encoder layers & 1 \\ 
        Attention hidden sizes & 32 \\ 
        Number of attention heads & 1 \\ 
        Attention drop out rate & 0 \\ 
        Feed forward layer hidden sizes & 32 \\ 
        Feed forward layer dropout rate & 0 \\ 
        Dense1 hidden sizes & 512 \\ 
        Dense1 dropout rate & 0.25 \\ 
        Dense2 hidden sizes & 1024 \\ 
        Dense2 dropout rate & 0.25 \\ 
        Dense3 hidden sizes & 288 \\ 
  \bottomrule
\end{tabular}
\end{table}

\subsection{Training}
We use a GTX 1080Ti for training our models, and the BERT model costs about 90 minutes. An RMSE loss function and Adam optimizer are used for training as shown in Table 2.

\begin{table}
  \caption{Training configuration}
  \label{tab:freq}
  \begin{tabular}{ccl}
    \toprule
    Training parameters&Value\\
    \midrule           
        Batch sizes & 1024 \\ 
        Training epochs & 3 \\ 
        Learning rate & 0.005 \\ 
        Optimizer & Adam \\ 
  \bottomrule
\end{tabular}
\end{table}

\subsection{Post-processing}
A series of post-processing strategies are employed to further improve forecasting accuracy. We know it's important for better prediction to capture the trend, seasonality, and spatial information. However, we find our forecasting results don't have obvious daily periodicity, while the descriptive analysis in historical data shows it's strong for most days. So we propose two ways to solve it. The first one is to add the daily fluctuation by post-processing. The daily average fluctuation is calculated and added to the prediction result directly. The second way is to optimize the model to make the model learn the daily fluctuation by itself, in which the time or wind information information is added as the decoder feature to decode the BERT output further. In the end, the latter methods in models are not as good as the first way in our experiments.

To give more details, we calculate the average wind power for every interval. Then the daily sequence is standardized to zero and one. Because the wind power is between 0 and 1620, a multiplier of 36 is chosen to magnify it according to local validation and the leaderboard. The multiplier could force the daily fluctuation to influence the predictions in a reasonable scope. For larger values, we further magnify them by multiplying a constant value like 1.1. Finally, the daily fluctuation sequence needs to match the predicted start time by shifting it, as not all predictions start from 00:00. In this way, the predicted future sequence could reflect the daily period with the same daily fluctuation level with historical data.

\section{Experiment}

\subsection{Validation strategy}
There are 3 phases of online test data in the competition, with a temporal relationship. So we leave a gap between offline training data and validation data to simulate this scenario. The training data we choose is between the 1st and 181st day, and the offline validation data is between the 231st and 245th day as shown in Figure1.

Considering we use 250982 valid samples as local validation, while the online test data use only 150-200 samples, we submit the model unless the local validation score is improved to decrease the risk of shaking down. Because the concept drift issue occurs for time-related task, in the last phase, we modified the model based on our best model in the second phase.

\subsection{Results and comparison}
In this section, we compare the different models' performances from our experiments. The results of the time series model and spatial-temporal model are shown in Table 3. To make the local validation comparable with the leaderboard, the MAE, RMSE and metrics scores are transferred to the sum of 195 samples' results.

\begin{table*}
  \caption{Performance comparison of models}
  \label{tab:commands}
  % 这个ccl害我纠结半天，原来需要改这里，非常尴尬
  \begin{tabular}{ccccc}
    \toprule
    Model & Local MAE &Local RMSE &Local score & Leaderboard score (phase II)\\
    \midrule
    BERT& 299 & 365 & 58.1 & 44.6  \\ 
    LSTM& 305 & 369 & 58.9 & 44.8  \\ 
    TCN & 310 & 371 & 59.4 & 45.1   \\ 
    KNN & 316 & 368 & 60.6 & unsubmit  \\
    LGB & 319 & 386 & 61.4 & unsubmit  \\ 
    Transformer & 311 & 374 & 60.0 & 48.5  \\ 
    Seq2seq & 296 & 366 & 57.6 & 47.1  \\ 
    Wavenet & 306 & 370 & 59.1 & 47.9   \\
    GCN-LSTM & 228 & 362 & 54.5 & 48.2  \\
  \bottomrule
\end{tabular}
\end{table*}

We can see that the deep learning methods behave better than statistical learning methods like KNN or LGB in our experiments. The wind power data is homogeneous, without important categorical information or crossed information, so the deep learning model can handle this task very well. The spatial information definitely could help more accurate predictions. As for the spatial relationship, we tried the spatial-temporal models to capture the spatial relationship automatically, but our implementations are not so successful in the leaderboard. Now that every turbine's power trend is quite similar, we focus more on its temporal properties afterward.

\section{Conclusion and future work}
In this technical report, we introduce our BERT solution for Baidu KDD Cup 2022 wind power forecasting. The BERT model can predict the primary trend for the long sequence prediction, and the daily fluctuation is added by post-processing to make the prediction with daily periodicity. Though the single BERT model is accurate and efficient, it can still be enhanced in many ways like transfer learning or the graph model.

% \section{Acknowledgements}
\begin{acks}
We would like to thank ACM and Baidu for hosting the KDD Cup 2022 Challenge. It’s a challenging task, bringing us an interesting journey. We also want to thank every competitive participant who makes our place keep dropping in leaderboard until last minute.
\end{acks}


\bibliographystyle{ACM-Reference-Format}
\bibliography{sample-base}

\end{document}